\documentclass{article}

\usepackage{amsmath,blkarray,bbold,graphicx}
\usepackage{float}
\usepackage{times}
\usepackage{graphicx} 
\usepackage{subfigure} 

\usepackage{natbib}

\usepackage{algorithm}
\usepackage{algorithmic}
\usepackage{amsmath,blkarray,bbold,graphicx}
\usepackage[colorinlistoftodos]{todonotes}
\usepackage{url}
\usepackage{amsmath}
\usepackage{times}
\usepackage{import}
\usepackage{natbib} %
\usepackage{hyperref} 
\usepackage{url} %
\usepackage{fancyvrb}
\usepackage{subfigure} %
\usepackage{comment} %
\usepackage{mdframed}
\usepackage{enumitem}
\usepackage{relsize}
\usepackage{framed}
\usepackage{graphicx} %
\usepackage{wrapfig}
\usepackage{bbm}
\usepackage[colorinlistoftodos]{todonotes}

\usepackage{multirow}

\usepackage[capitalize]{cleveref}  

\crefname{nlem}{Lemma}{Lemmas}
\crefname{nprop}{Proposition}{Propositions}
\crefname{ncor}{Corollary}{Corollaries}
\crefname{nthm}{Theorem}{Theorems}
\crefname{assumption}{Assumption}{Assumptions}

\usepackage{misc}

\usepackage{hyperref}

\usepackage[accepted]{icml2016} 

\icmltitlerunning{The Segmented iHMM}

\begin{document} 

\twocolumn[
\icmltitle{The Segmented iHMM: A Simple, Efficient Hierarchical Infinite HMM}

\icmlauthor{Ardavan Saeedi}{ardavans@mit.edu}
\icmladdress{Computer Science and Artificial Intelligence Laboratory,
MIT}
\icmlauthor{Matthew Hoffman}{mathoffm@adobe.com}
\icmladdress{Adobe Research}
\icmlauthor{Matthew Johnson}{mattjj@csail.mit.edu}
\icmladdress{Harvard University}
\icmlauthor{Ryan Adams}{rpa@seas.harvard.edu}
\icmladdress{Harvard University and Twitter}

\vskip 0.2in
]

\begin{abstract}
We propose the segmented iHMM (siHMM), a hierarchical infinite hidden Markov model (iHMM) that supports a simple, efficient inference scheme. The siHMM is well suited to segmentation problems, where the goal is to identify points at which a time series transitions from one relatively stable regime to a new regime. Conventional iHMMs often struggle with such problems, since they have no mechanism for distinguishing between high- and low-level dynamics. Hierarchical HMMs (HHMMs) can do better, but they require much more complex and expensive inference algorithms. The siHMM retains the simplicity and efficiency of the iHMM, but outperforms it on a variety of segmentation problems, achieving performance that matches or exceeds that of a more complicated HHMM.
\end{abstract}

\section{Introduction}

\begin{figure*}[t!]
\begin{center}
\begin{tabular}{cc}
{\includegraphics[trim = 0mm 0mm 10mm 0mm, clip, scale = 0.31]{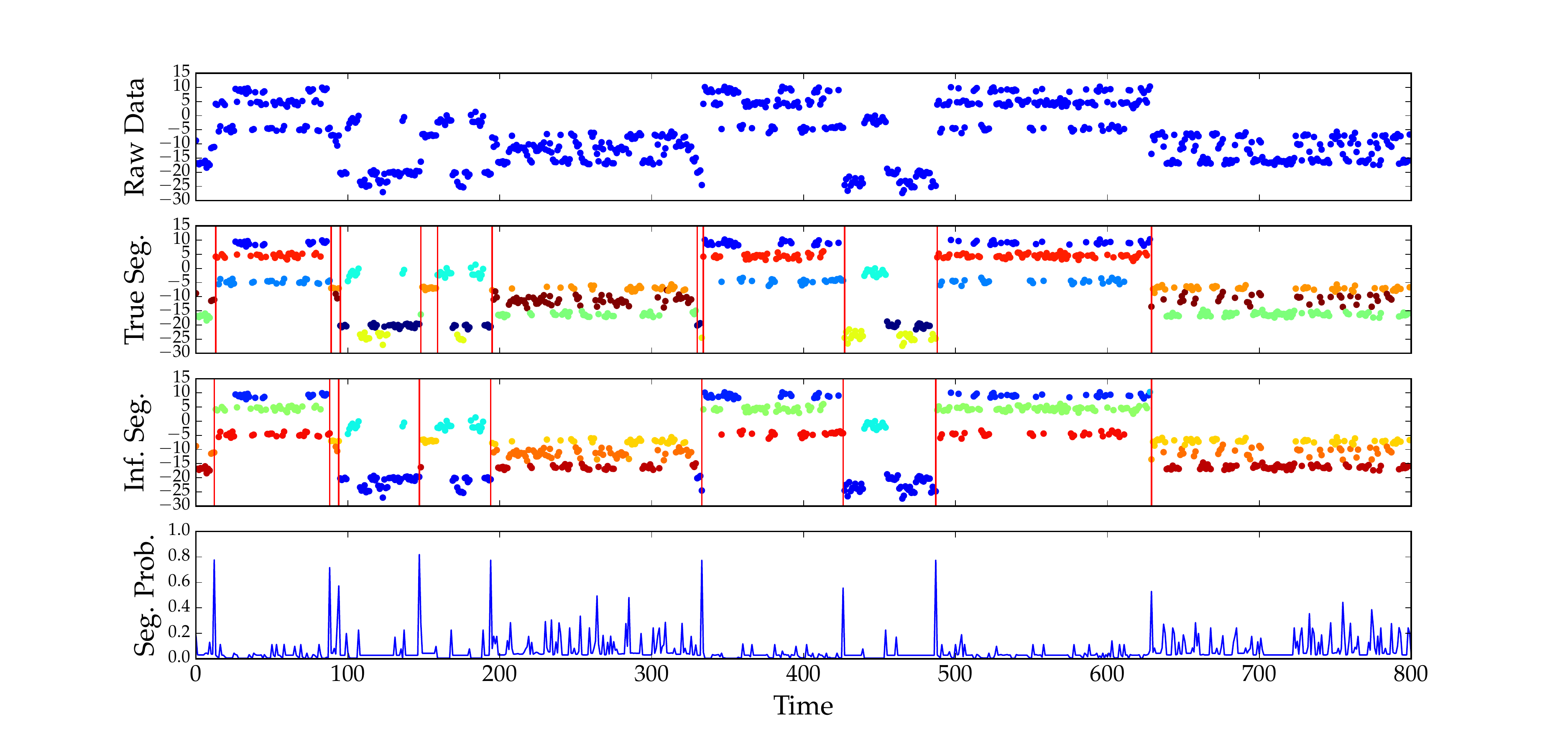}} &  
 \begin{tabular}[b]{c}\hspace{-1.2cm}\includegraphics[trim = 30mm 5mm 0mm 0mm, clip, scale = 0.26]{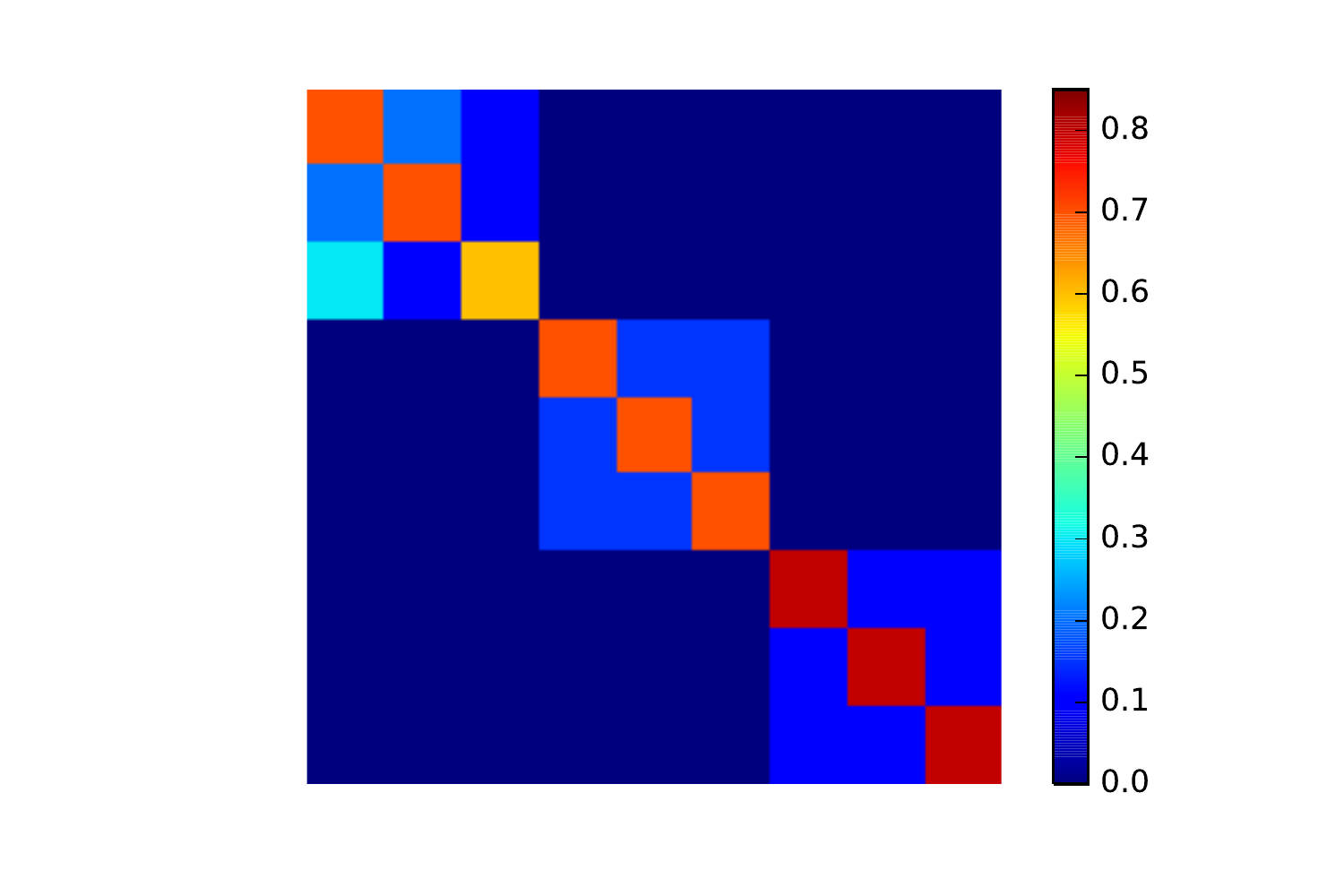} \\ \vspace{0.1cm}\hspace{-1.7cm}\includegraphics[trim = 0mm 0mm 0mm 0mm, clip, scale = 0.16]{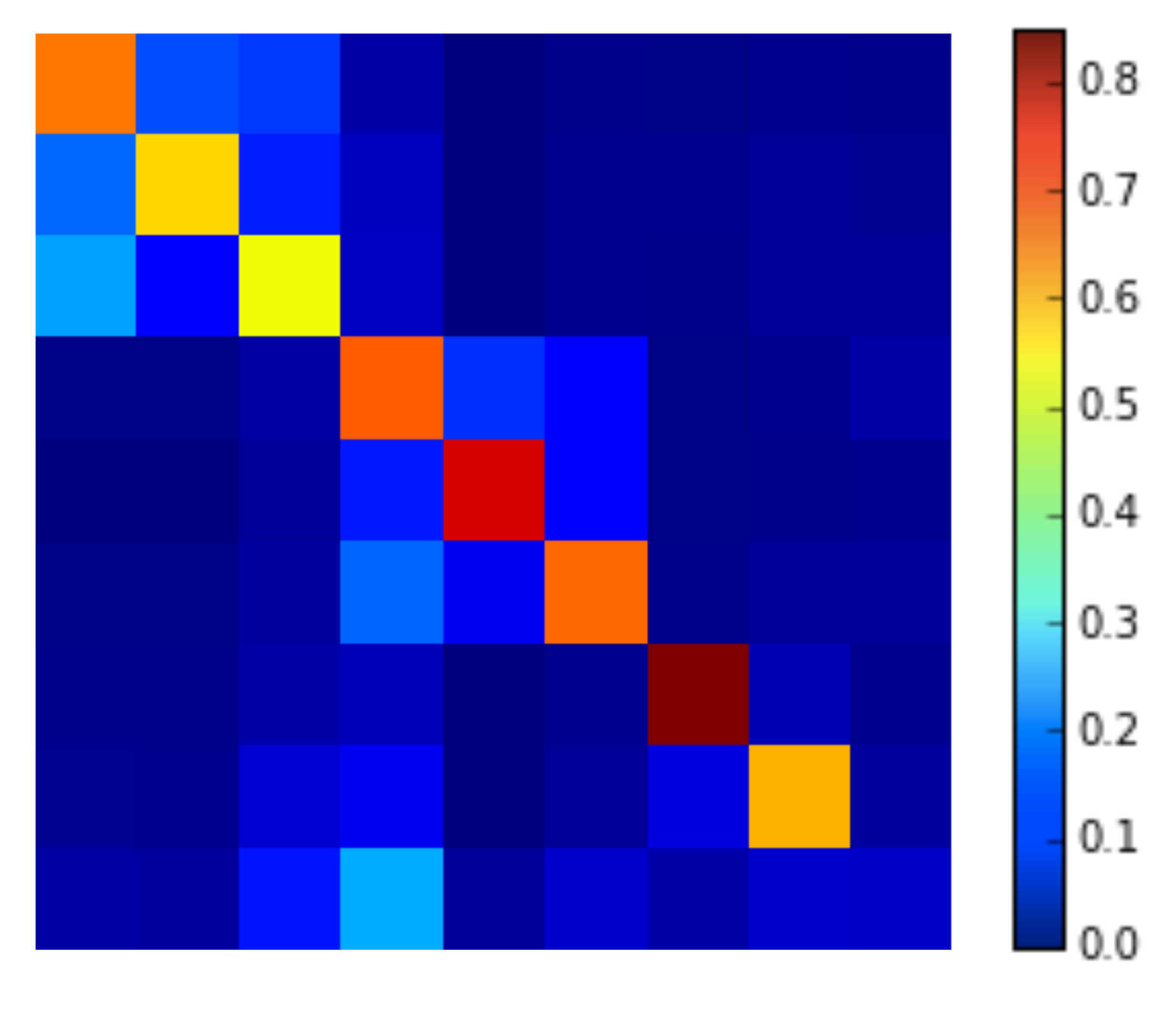}\end{tabular} \\
\footnotesize{(a)} & \hspace{-2cm}\footnotesize{(b)} \\ 
\end{tabular}
\end{center}
\caption{\footnotesize{(a) A sample of a synthetic dataset with true and inferred segmentations. Top three plots: the y-axis shows 1-d observations, colors denote true or inferred hidden state, vertical lines denote true or inferred segment boundaries. Bottom plot: inferred posterior probability of a segment boundary. (b) \textbf{Top} True and \textbf{Bottom} inferred transition matrices}}
\label{fig:toydata}
\end{figure*}

The infinite hidden Markov model (iHMM)~\cite{beal2001infinite} and its variants (e.g., \cite{fox2008hdp, saeedi2011priors}) have been among the most successful Bayesian nonparametric models, with applications from speech recognition~\cite{fox2011sticky} to biology~\cite{beal2012gene}. However, despite their success in modeling time series with complicated low-level dynamics, their application to time series with multiple timescales has been limited.

Such hierarchically structured sequences characterized by multiple timescales arise in many domains, such as natural language~\cite{lee2013joint}, handwriting \cite{lake2014one}, and motion recognition \cite{heller2009infinite}. For example, it is natural in motion recognition to model the sequence of high-level actions (such as walking to a chair, then sitting down) and steps within the actions (e.g., bending one's knees then leaning back to sit down) at two different levels.

We will focus on the problem of \emph{segmentation}, in which the goal is to identify points at which a time series transitions from one relatively stable regime to a new regime. In the motion recognition example, the segmentation problem would be to identify when a subject transitioned from one type of action (e.g., walking) to another (e.g., sitting down), without necessarily identifying what they were doing. This is one of the easiest problems in time-series modeling that involves multiple timescales, but (as we will see) it is quite hard for (i)HMMs, which have no mechanism for distinguishing between high- and low-level dynamics.

The hierarchical HMM (HHMM) is a generalization of the HMM that naturally deals with dynamics at multiple timescales \cite{fine1998hierarchical, murphy2002linear}. But this generality comes at a price: these models lack the simple predictive distributions and efficient inference schemes that make (i)HMMs so popular.  And the available nonparametric versions of the HHMM such (e.g., \cite{heller2009infinite, stepleton2009block}) are complex to implement and not readily amenable to efficient inference. 

In this paper, we propose the segmented iHMM (siHMM), a simple extension to the iHMM that does not suffer from the above problems and can discover segment boundaries in time series with dynamics at two timescales. Unlike the HHMM, our model does not explicitly model higher-level state; instead, it assumes dynamics that evolve according to a standard iHMM except for occasional change-point events that kick the model into a new randomly chosen hidden state, disrupting the low-level dynamics of the iHMM. Because it relies on a very simple model of high-level dynamics, inference in the siHMM has time and implementation complexity similar to that of the iHMM, and well below that of a typical HHMM.  We show that this simple change-point extension is sufficient to encourage the iHMM to model time-series data characterized by multiple regimes of low-level dynamics.
Although our model is limited by the depth of the hierarchy, in many practical applications of HHMMs (e.g., \cite{olivera2004layered, nguyen2005learning, xie2003unsupervised}) a two-level analysis of the dynamics is sufficient.


Below, we describe two versions of the model. The first version, which we call the feature-independent model, enjoys conditional conjugacy and therefore has simple Gibbs and variational inference algorithms. The second version, which we call the feature-based model, can incorporate domain knowledge without requiring a complex new machinery. We present an stochastic variational inference (SVI) scheme for the feature-based model; the derivation for the feature-independent model is similar and straightforward.


We apply the model to three different tasks: a novel task of segmenting traces of user behavior in software applications, automatic behavioral segmentation of fruit fly and sensor data labeling. Segmenting user behavior traces is of significant importance in understanding the behavior of software application users; it can help in identifying and simplifying the complex common patterns among the users (e.g., \citet{adar2014commandspace, han2007frequent, horvitz1998lumiere}). For the fruit fly behavior segmentation, we use a dataset from \citet{kain2013leg}; the results of this task can be used to better understand how the nervous system generates behavior. Finally, labeling sensor data gathered in everyday life settings can be used not only to understand physical activities (e.g., \citet{ermes2008detection,parkka2006activity}), but also to detect psychological and emotional states (e.g., \citet{picard2001toward,healey2005detecting}). Implementing effective health and wellbeing related interventions, understanding user behavior, and designing affective interfaces, are only a few applications of this task.

We empirically compare our model with two main baselines: 1) a two-level Bayesian nonparametric hierarchical HMM (HHMM) introduced by  \citet{johnson2014bayesian} that models high-level dynamics as an infinite hidden semi-Markov model (HSMM) and sub-dynamics as an iHMM, and 2) the iHMM. In each of these tasks, we show that our model outperforms the nonparametric  HHMM (despite being substantially simpler and faster) and the iHMM.

\section{Model}
\label{sec:model}


Our model can be viewed as a generalization of an iHMM where the transition probability from each state $k$ is a mixture of two distributions: 1) a state-dependent transition probability distribution $\pi_k$, as in an iHMM, and 2) a state-independent probability distribution $\pi_0$. Which of these two distributions generates a hidden state $z_t$ at a given time $t$ depends on the hidden state $z_{t-1}$ and observation $y_{t-1}$ at the previous time $t-1$.

We say that the transitions caused by $\pi_0$ define the boundaries of a \emph{segment}. The model implicitly assumes that the low-level dynamics within a segment are more structured and predictable than the higher-level dynamics that govern transitions between segments, since it can throw much more modeling power at these low-level dynamics. In motion capture data, for instance, the dynamics of a walk may be highly structured and predictable, whereas the dynamics that govern whether a user transitions from walking to standing, sitting, or running may be much less predictable. 


\subsection{Feature-independent model} 

The feature-independent model assumes the following generative process. At time step $t = 0$, we initialize the process by sampling a hidden state $z_0$ from a distribution $\pi_0$. Given a hidden state $z_t$, we generate an observation $y_t$ from a conditional observation distribution $F(\phi_{z_t})$ where $\phi_{z_t}$ is the parameter corresponding to the hidden state $z_t$: $y_t|z_t \sim F(\phi_{z_t})$.

\begin{figure}[bt]
\begin{center}
\includegraphics[width=0.8\columnwidth ]{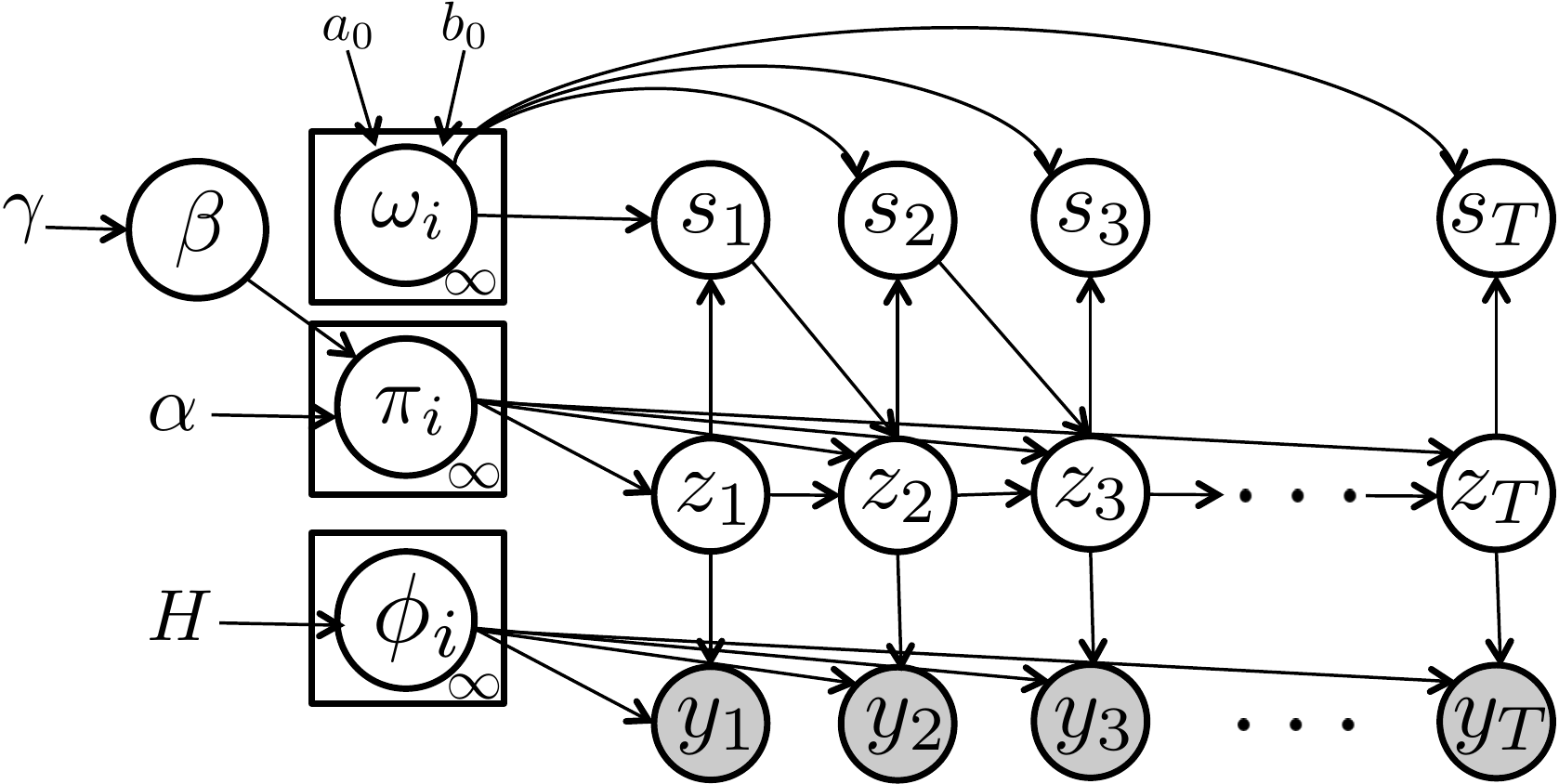} 
\caption{\footnotesize{Graphical representation of the feature-independent siHMM.}}
\label{fig:model1}
\end{center}
\end{figure}

Next, we sample a variable $s_t$, which we call the segmentation variable, from a Bernoulli distribution with a parameter $\omega_{z_t}$. This is a state-dependent variable which has a conjugate beta prior with hyperparameters $a_0$ and $b_0$.  Here, $s_t = 1$ denotes the beginning of a new segment: 
\[
\nonumber
\omega_i \sim \text{Beta}(a_0, b_0); &\quad s_t|z_t,y_t \sim \text{Bern}(\omega_{z_t}).
\]
We denote the probability of creating a new segment at time $t$ by $p^{\text{seg}}_t$.
If $s_t = 0$, we sample the next state $z_{t+1}$ from a state-dependent distribution $\pi_{z_t}$ (as in the iHMM), otherwise, we ignore the current state $z_t$ and sample $z_{t+1}$ from a distribution $\pi_0$:
\[
\nonumber
z_{t+1}\mid z_t, s_t = 0 \sim \pi_{z_t},
\quad
z_{t+1}\mid z_t, s_t = 1 \sim \pi_0.
\]
The transition matrix has the same generative process as the iHMM:
\[
\nonumber
\beta \sim \text{GEM}(\gamma);\qquad
&\pi_i \sim \text{DP}(\alpha\beta);\qquad
\phi_i \sim H;
\]

where $H$ is the prior distribution over $\phi$, GEM$(\gamma)$ is the stick-breaking distribution with concentration parameter $\gamma$, and DP$(\alpha \beta)$ denotes sampling from a Dirichlet process with concentration parameter $\alpha$. The graphical model of the feature-independent siHMM is depicted in Fig. \ref{fig:model1}.

An illustration of the model applied to a synthetic dataset (explained in Sec.~\ref{sec:experiments}) is provided in Fig.~\ref{fig:toydata}. The model is able to approximately recover the block-diagonal structure of the true transition matrix. Even though the model does not explicitly encourage block-diagonal structure, the sparsity induced by the DP prior on $\pi_k$ is sufficient to encourage the model to push inter-segment dynamics into $\pi_0$ and recover the block-diagonal intra-segment dynamics.

\subsection{Feature-based model}
\label{sec:Feature-based model}
In some tasks such as segmenting software user traces or tagging fruit fly behavior, there is a rich domain knowledge available for improving the model. For instance, in segmenting user traces features like $\mathbbm{1}(\text{action} = \texttt{save})$ or $\mathbbm{1}(\text{action} = \texttt{close})$ may indicate the end of a segment. We modify the model in a way that we can add features declaratively. Although due to lack of conjugacy, deriving the Gibbs sampler is not straightforward anymore, in Section~\ref{sec:inference}, we derive an efficient SVI algorithm for this model. 

\begin{figure}[bt]
\begin{center}
\includegraphics[width=0.8\columnwidth ]{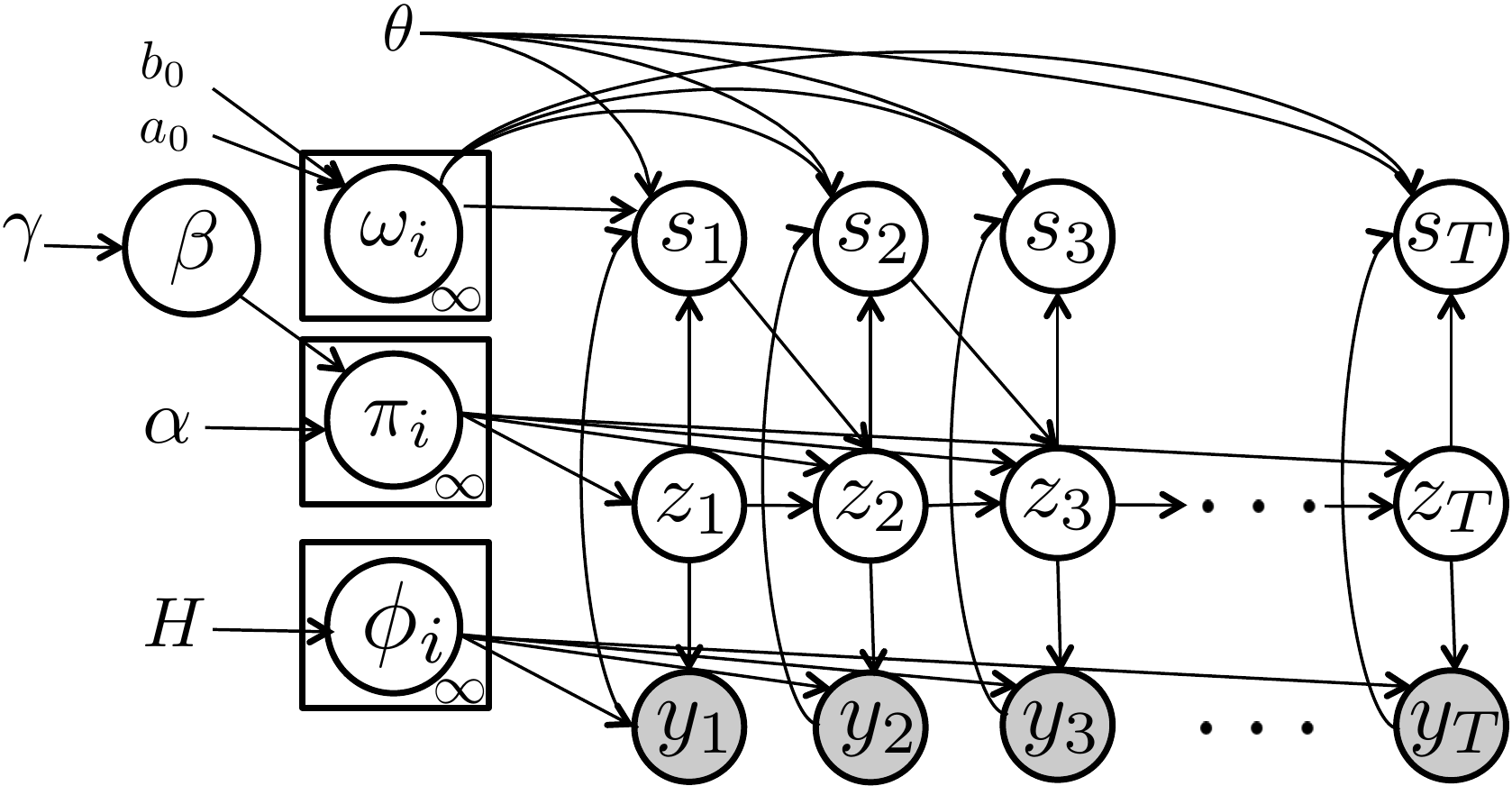} 
\caption{\footnotesize{Graphical representation of the feature-based siHMM.}}
\label{fig:model2}
\end{center}
\end{figure}

The difference between this version of the model and the feature-independent version is in the conditional distribution of the segmentation variable $s_t$ (see Fig. \ref{fig:model2} for the graphical model). Here, the parameter of the Bernoulli distribution is 
\[
\nonumber
\sigma(\theta \cdot \text{f}(y) + \omega_{z_t}) = \frac{\exp{(-\theta \cdot \text{f}(y) - \omega_{z_t})}}{1+\exp{(-\theta \cdot \text{f}(y) - \omega_{z_t})}}
\]
where $\theta$ is the weight vector for the data-dependent features, $\text{f}(y)$ is the feature function which consists of all observation-dependent features, and $\omega_{z_t}$ is the feature weight for hidden state $z_t$. To simplify the notation, we assume that the observation-dependent features only depend on the observation at a single time step. Hence, we have 
\[
\nonumber
s_t|z_t,y_t \sim \text{Bern}(\sigma(\theta \cdot \text{f}(y_t) + \omega_{z_t})).
\]
We do not assume a prior for the feature weights; instead, we use a point estimate for them in our SVI algorithm.  

\section{Stochastic variational inference}
\label{sec:inference}
To keep the notation uncluttered, we assume that we have a dataset $\bf{y}$ of $K$ sequences all with the same length $T$ and write: $\textbf{y} = \{y^{k}_{1:T}\}^K_{k=1}$, $\textbf{z} = \{z^{k}_{1:T}\}^K_{k=1}$, $\textbf{s} = \{s^k_{1:T}\}_{k=1}^K$. For inference, we use the stochastic variational inference (SVI) algorithm \cite{hoffman2013stochastic} and approximate the posterior with a truncated variational distribution introduced in \cite{johnson2014stochastic}. We approximate the posterior $p(\textbf{z}, \textbf{s}, \beta, \theta, \omega, \pi, \phi|\textbf{y})$ with mean field family distribution $q(\textbf{z} ,\textbf{s})q(\beta)q(\omega)q(\theta)q(\pi)q(\phi)$. In the language of SVI, $\textbf{z}$ and $\textbf{s}$ are local variables and $\beta$, $\omega$, $\theta$, $\pi$, and $\phi$ are global variables. We maximize the marginal likelihood lower bound $\mathcal{L}$: 
\begin{equation}
\nonumber
\mathcal{L} \triangleq \mathbb{E}_{q}\left[\frac{p(\textbf{z}, \textbf{s}, \beta, \omega, \theta, \pi, \phi,\textbf{y})}{q(\textbf{z}, \textbf{s})q(\beta)q(\omega)q(\theta)q(\pi)q(\phi)} \right]
\end{equation}
by using stochastic natural gradient ascent over the global factors and standard mean field updates for the local factors. At each iteration of SVI, we sample a minibatch of $M$ sequences from the dataset and update its local factors; next, given the expectation with respect to the local factors, we update the global factors by taking a step of size $\rho$ in the approximate natural gradient direction. To further simplify the notation, we assume that the minibatch is a single sequence and drop the superscript for $y$, $z$ and $s$. Next, we explain the variational factors for each of the variables.  

\subsection{Variational factors}
For $q(z_{1:T}, s_{1:T})$, the ``direct assignment'' truncation used in \cite{johnson2014stochastic}, sets $q(z_{1:T}, s_{1:T}) = 0$ if for any of $z_1$ to $z_T$ we have $z_t = k$ and $k > K$; here $K$ is the truncation level. Since by using this truncation the update to $q(\beta)$ conditioned on the other factors is not conjugate anymore, we use a point estimate for $q(\beta)$: $q(\beta) = \delta_{\beta^*}(\beta)$. We adopt the same point estimate approach for the parameters of the sigmoid function; hence, $q(\theta)=\delta_{\theta^*}(\theta)$ and  $q(\omega_z)=\delta_{\omega_z^*}(\omega_z)$.

With this truncation scheme, we can write the prior over $\pi_i$ as $p((\pi_{i1}, \dots, \pi_{iK}, \pi_{i, \text{rest}})) = \text{Dir}(\alpha \beta_1, \dots, \alpha \beta_K, \alpha \beta_{\text{rest}} )$. Here, $\pi_{i,\text{rest}} = 1 - \sum_{k=1}^K \pi_k$ and $\beta_{\text{rest}} = 1 - \sum_{k=1}^K\beta_k$. We know from \cite{hoffman2013stochastic} that due to conjugacy the optimal $q((\pi_{i1}, \dots, \pi_{iK}, \pi_{i, \text{rest}}))$ is in the form of $\text{Dir}(\tilde{\alpha}_i)$ where $\tilde{\alpha}$ is the parameter of the variational distribution.

We assume that the prior over $\phi$ is in exponential family with natural parameter $\eta$, and it is a conjugate prior for the likelihood function $f(y_t|\phi)$. This implies that the optimal variational distribution $q(\phi)$ is also in the same family with some other natural parameter denoted by $\tilde{\eta}$. More formally, we have: $h(\phi_i) \propto \exp\{\langle \eta_i, t_{\phi}(\phi_i) \rangle\}$ and $q(\phi_i) \propto \exp\{\langle \tilde{\eta}_i, t_{\phi}(\phi_i) \rangle\}$ where $t_{\phi}$ is the sufficient statistic function of $p(\phi)$.

\subsection{SVI update equations}
\label{sec:sviupdate}
For the variational updates, we need to take expectations with respect to each of the variational distributions. For the expectations with respect to $q(z_{1:T}, s_{1:T})$, a modification of the standard HMM forward-backward algorithm with the following forward $F$ and backward $B$ messages can be used:

\vspace{-2mm}
\[
\label{eq:fb}
\nonumber
F(z_t &, s_t) \triangleq f(y_t|\phi_{z_t})p(s_t|z_t,y_t) \times\\ 
& \sum_{z_{t-1},s_{t-1}}F(z_{t-1},s_{t-1})p(z_t|s_{t-1},z_{t-1});\\
\nonumber
B(z_t &, s_t) \triangleq \sum_{z_{t+1},s_{t+1}}B(z_{t+1},s_{t+1}) \times \\ 
&f(y_{t+1}|\phi_{z_{t+1}})p(s_{t+1}|z_{t+1},y_{t+1})p(z_{t+1}|s_t,z_t).
\]
 These messages can be computed in $O(TK^2)$. In fact, the augmented transition matrix that we need to compute for these forward-backward messages has the following form:
\[
\vspace{-100mm}
\nonumber
\begin{blockarray}{cccc}
        & s_{t+1}=0 & & s_{t+1}=1 \\
      \begin{block}{c[ccc]}
        s_t=0 & (1-p_{t+1}^{\text{seg}})\Pi & | & p_{t+1}^{\text{seg}}\Pi \\
        & -------- && -------- \\
        s_t=1 & (1-p_{t+1}^{\text{seg}}) \scalebox{1.2}{$\mathbf{1}$} \pi_0^T & | & p_{t+1}^{\text{seg}} \scalebox{1.2}{$\mathbf{1}$} \pi_0^T\\
      \end{block}
\end{blockarray}
\]
where $\Pi$ is a $K \times K$ transition matrix and $\scalebox{1.2}{$\mathbf{1}$}$ is an all-ones vector of size $K$. The matrix operation for computing a message requires $2K^2$ operations for the upper half of the matrix ($s_t = 0$) and $2K + K$ for the lower half ($s_t=1$). This is because all the rows of each $K \times K$ block in the lower half are the same. Hence, the total number of operations for message passing is $T(2K^2+3K)$. Furthermore, the total memory required to compute these operations is $O(TK)$. 

For updating the local factors, instead of $\pi_{ij}$ and $f(y_t|\phi_{z_t})$ in Eq.\ref{eq:fb}, we compute the forward-backward messages using: $\tilde{\pi}_{ij} \triangleq \exp\{\mathbb{E}_{q(\pi)}\ln \pi_{ij}\}$ and $\tilde{L}_{ti} \triangleq \exp\{\mathbb{E}_{q(\phi_i)}\ln f(y_t|\phi_{z_t}, z_t = i)\}$

The expectations of the sufficient statistics with respect to $q(z_{1:T}, s_{1:T})$ are:
\vspace{-2mm}
\[
\nonumber
\tilde{t}^{ij}_{\text{trans}} &\triangleq \mathbb{E}_{q(z_{1:T}, s_{1:T})}\sum_{t=2}^T \mathbbm{1}[z_{t-1} = i, z_{t} = j, s_{t-1}=1, s_{t} = .];\\
\nonumber
\tilde{t}^{0j}_{\text{trans}} &\triangleq \mathbb{E}_{q(z_{1:T}, s_{1:T})}\left[\mathbbm{1}[z_1 =j] + \sum_{t=2}^T \mathbbm{1}[z_t = j, s_{t-1} = 1] \right]; \\
\tilde{t}^j_{y} &\triangleq \mathbb{E}_{q(z_{1:T}, s_{1:T})}\sum^T_{t=1}\mathbbm{1}[z_t = j, s_t = .]t_y^j(y_t).
\]
Given these expected sufficient statistics and a scaling factor $m \triangleq S / M$, we can write the update equations for the parameters of the global variational factors $q(\pi)$ and $q(\phi)$:
\vspace{-1mm}
\begin{equation}
\begin{split}
\nonumber
\tilde{\eta}_i &\leftarrow (1-\rho) \tilde{\eta}_i + \rho (\eta_i + m.\tilde{t}^i_y) \\
\tilde{\alpha}_i &\leftarrow (1-\rho) \tilde{\alpha}_i + \rho (\alpha_i + m.\tilde{t}^{i}_{\text{trans}})\\
\tilde{\alpha}_0 &\leftarrow (1-\rho) \tilde{\alpha}_0 + \rho (\alpha_0 + m.\tilde{t}^{0}_{\text{trans}}).
\end{split}
\end{equation}
For the global factors $q(\omega)$, $q(\theta)$ and $q(\beta)$, we use a point estimate; hence, we only need the gradient of $\mathcal{L}$ with respect to $\beta^*$, $\theta^*$ and $\omega^*$. For $\nabla_{\beta^*}$ we follow the derivation in \cite{johnson2014stochastic} and obtain the following gradient to use in a truncated gradient step on $\beta^*$:
\[
\nonumber
\nabla_{\beta^*} \mathcal{L}  &= \nabla_{\beta^*} \left\lbrace \mathbb{E}_{q(\pi)}\left[ \ln \frac{p(\beta, \pi)}{q(\beta)q(\pi)} \right] \right\rbrace \\
\nonumber
&=\nabla_{\beta^*} \left\lbrace \ln p(\beta^*) + \sum^K_{i=1}\mathbb{E}_{q(\pi_i)}\ln p(\pi_i|\beta^*) \right\rbrace
\]
Note that, we need to ensure that $\beta^* \geq 0$ after each update. To estimate $\theta^*$, we have:
\[
\nonumber
\nabla_{\theta^*} \mathcal{L} &= \nabla_{\theta^*} \left\lbrace \mathbb{E}_{q(z_{1:T}s_{1:T})}\left[ \ln \frac{p(\theta, z_{1:T}, s_{1:T}, y_{1:T})}{q(\theta)q(z_{1:T}s_{1:T})} \right]  \right\rbrace \\
\nonumber
= \mathbb{E}&_{q(z_{1:T},s_{1:T})}\left[ \sum^{T}_{t=1}s_t \text{f}(y_t) - p(s_t=1|y_t,z_t)\text{f}(y_t)\right].
\]
We have a similar equation for $\omega^*_z$. 
  
\section{Related models}
\label{sec:relatedmodels}
There are few models similar to our model in terms of extending iHMM to multiple timescales. Infinite hierarchical HMM (iHHMM), introduced in \cite{heller2009infinite}, is a nonparametric model that allows the HHMM to have potentially unbounded depth. Hence, the model can infer the number of levels in the  hierarchy. In iHHMM, the bottom level is the observed sequence and each level is a sequence of hidden variables dependent on the level above. As the authors suggested, more efficient inference algorithms are needed in order to make their model useful for practical applications.  

The block-diagonal iHMM \cite{stepleton2009block} is a generalization of iHMM that assumes a nearly block-diagonal structure on the transition matrix of the iHMM. Each block corresponds to a ``sub-behavior'' and the model can partition the data sequences according to these sub-behaviors. The model first partitions the infinite number of hidden states into an infinite number of blocks by using an additional stick-breaking process. Then, it increases the probability of transition between the states of a block by modifying the Dirichlet process prior over the transitions. Hence, as the block size becomes smaller, the model behavior converges to that of iHMM. For inference, as the authors explained, achieving a fast mixing rate in their proposed inference algorithm requires implementing a nontrivial bookkeeping-intensive method. In contrast, our model is much simpler and easier to implement inference for, but it can also discover transition matrices with approximately block-diagonal structure; the segmentation events provide a mechanism for transitioning from one group of connected states to another. 

Another related model is a two-level Bayesian nonparametric HMM introduced in~\cite{johnson2014bayesian} that models the high-level dynamics or the superstates as an HDP-HSMM. As a generalization of iHMM, HDP-HSMM can model the dwell time in each state by sampling that from a state-specific duration distribution once a state is entered. For a formal definition of the HDP-HSMM, see \cite{johnson2014bayesian}.   
Given each superstate $j$, observations are generated according to an iHMM with parameters $\{\beta^j, \pi^{j}_k, \phi^j_k \}^{\infty}_{k=1}$.
where $x_t$ denotes the substate at time step $t$.  Compared to our model, this model is much more flexible; however, the computation of forward-backward messages is less efficient. Moreover, it requires more bookkeeping for all superstates and their substates. Compared to our model, it needs setting a truncation level for the superstates and each of the iHMMs correpsonding to them. We can set a large truncation level for all of them, but this means we are paying a huge computational cost for iHMMs that only require few states. In contrast, we only need to set a single truncation level for the whole model. We call this model sub-iHMM and use it as a baseline since it is a flexible Bayesian nonparametric model that supports two-level dynamics.

Finally, our model can also be applied to time-series segmentation tasks at a single level. The iHMM or a variant of it with self-transition bias, sticky HDP-HMM  \cite{fox2011sticky}, can be used for the same purpose. The self-transition bias encourages self-transitions, and consecutive hidden states tend to belong to one state. The model can capture segments in tasks such as speaker diarization; however, in contrast to our model, within a given state there are no dynamics. In other words, in the sticky HDP-HMM, given a state, the observations are independent of each other. This makes the model inappropriate for tasks such as user trace segmentation; that is because there has to be some order on substates within a segment. For instance, in a software application an action like \texttt{selection} needs to come before an action like \texttt{move selection}. To show the importance of this point, we also include a self-transition bias term in the iHMM and compare our model with sticky HDP-HMM \footnote{We do not include a separate column in Table~\ref{tbl:experiments} for sticky HPD-HMM; instead, we find the best variational lower bound for all settings of hyperparameters of iHMM including a hyperparameter for self-transition bias.}.


\section{Experiments}
\label{sec:allexperiments}
We evaluate the performance of our feature-independent and feature-based models on synthetic and real datasets.  We use a synthetic dataset to illustrate the advantages of our model compared to baselines.  

In order to further demonstrate the capabilities of our model, we apply it to three real tasks from three different domains: human-computer interaction, biology, and sensor data analysis. As mentioned in Section~\ref{sec:relatedmodels}, two reasonable baselines for our model are the two-level Bayesian nonparametric HMM and iHMM. We report the labeling error (or normalized Hamming distance in the case of the synthetic dataset) and predictive log-likelihood for our model and these baselines. To choose among different hyperparameter settings, we use the variational lower bound (VLB) as our objective measure. We show that our model, while being simpler and efficient in terms of inference, is competitive with or outperforms these baselines. For all experiments on siHMM we try both the feature-independent and feature-based models; we report the results separately in Table \ref{tbl:experiments} to show the effect of including observation features in the model. In the experiments, we only try the hidden state and the observation as the features for a given time step. However, more sophisticated features can be made from the observation(s). 



\subsection{Synthetic data}
\label{sec:experiments}
We generate a synthetic dataset with 5000 data points from 3 different transition matrices, each with 3 hidden states. Each row of each transition matrix is sampled from a modified $\distDir(1)$ with self-transition bias of 1. The observations are sampled from normal distributions with non-conjugate separate priors $\distNorm(0, 10)$ and $\distInvGam(2, 1)$ on their mean and variance parameters, respectively. The goal is to find the points where we change regimes and also to determine the dynamics within each segment of the sequence. At each time step with probability 0.05, we switch the regime. 

Fig.~\ref{fig:toydata} shows a sample sequence and the result of running 100 passes of SVI over the whole dataset for that sequence. For running SVI, we split the dataset into 20 sequences and use a batch of size 2. We randomly sample a sequence with length 750 to calculate the predictive log-likelihood. We report the error over the dataset for the hyperparameter setting with the highest VLB. 

For the hyperparameters in the feature-independent model, we set $a_0 = 1$ and $b_0 = 1$ for the beta prior; for the feature-based model, we randomly initialize the feature weights from $\distNorm(0, 1)$ and only use the hidden state as a feature. We do a grid search over combinations of values for $\alpha = \{1, 5\}$, $\gamma = \{1, 5\}$, and $K = \{20, 30\}$. We place an $\distNorm(0, 10)$ prior on the mean of the observation distributions. Their variance prior is $\distInvGam(\alpha_{g}, \beta_{g})$ where $\alpha_{g}$ and $\beta_{g}$ are coming from $\{1, 10\}$ and $\{0.1, 1\}$. We run SVI with 10 different seeds for 100 iterations over all these combinations. To compute the normalized Hamming distance between the inferred states and the true states, we use the Munkres algorithm \cite{munkres1957algorithms}. The algorithm assigns indices to the inferred sequence so that it maximizes the overlap with the true sequence. Table~\ref{tbl:experiments} gives the computed distance over the dataset for the hyperparameter setting with the highest VLB. We also report the predictive log-likelihood over the held-out set.  

In addition to the above hyperparameters, the sub-iHMM also requires the truncation level of the superstates and the substates, which we set to 10 and $\{5,10\}$, correspondingly. For all hyperparameters, shared with the siHMM, we use the same set of settings for the baselines. For the self-transition bias for sticky HDP-HMM we try $\kappa \in \{1,10, 100\}$. 

\begin{figure}[bt]
\begin{center}
\includegraphics[width=1.0\columnwidth ]{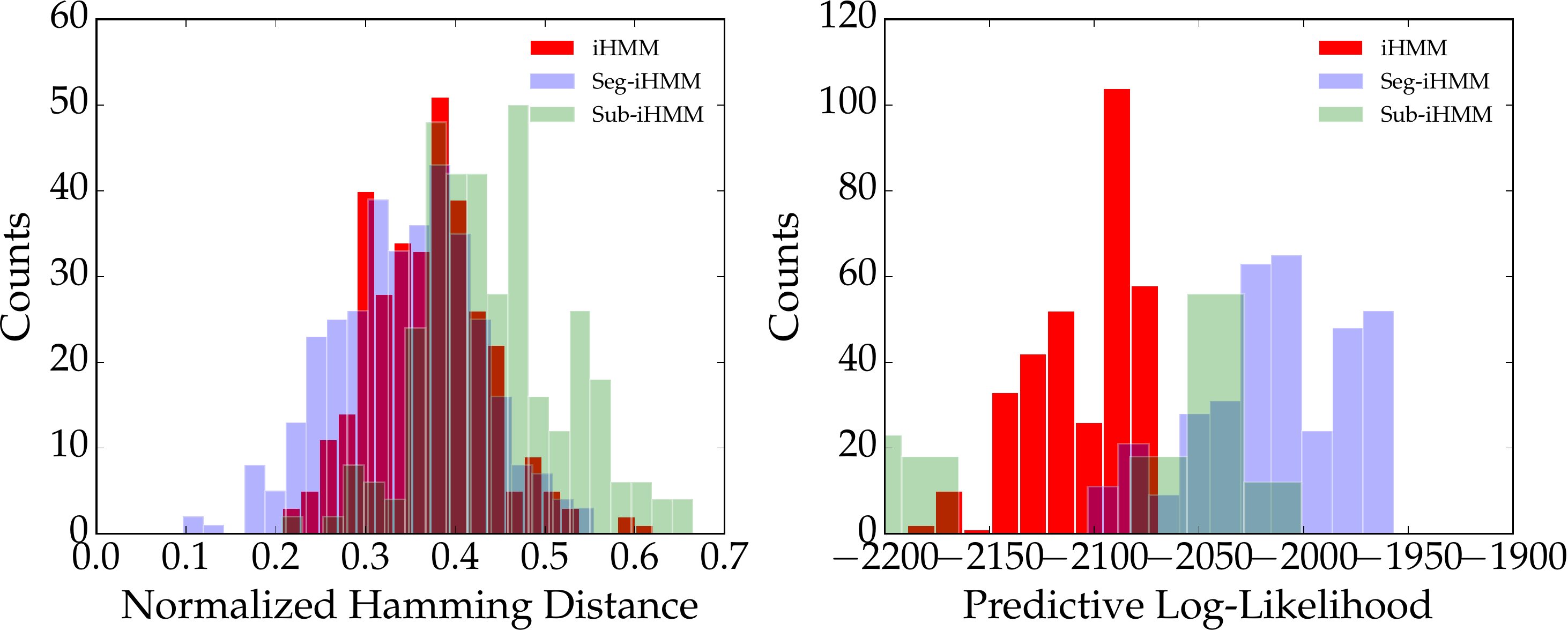} 
\vspace{-3mm}
\caption{\footnotesize{Histogram of normalized Hamming distance and predictive log-likelihood for different settings of hyperparameters for the synthetic data set.}}
\label{fig:synthhist}
\end{center}
\vspace{-8mm}
\end{figure}

Fig.~\ref{fig:synthhist} shows the histogram of the normalized Hamming distance and also the predictive log-likelihood for runs with different hyperparameter settings. In terms of the Hamming distance, the siHMM performs slightly better than the iHMM and outperforms the sub-iHMM for most settings. The same conclusion holds for the predictive log-likelihood. Furthermore, Fig.~\ref{fig:toydata} shows that our model can do reasonably well in finding the regime change points and also the states within each segment of the sequence. We choose a threshold of 0.5 for the posterior segmentation probability to identify a time point as a change point; however, as shown in the last row of Fig.~\ref{fig:toydata}, the model is able to provide an estimate for the uncertainty over change points. 

Table~\ref{tbl:experiments} shows that for the dataset which we generated from a block-diagonal transition matrix, siHMM outperforms both iHMM and sub-iHMM. This might be because the synthetic dataset is specifically generated without any state-dependent high-level dynamics. Our model, which does not assume any dynamics for the segments, performs better than the other baselines which implicitly or explicitly assume that.

%

\begin{table}[h]
\caption{Datasets used for experiments (description in text)}
\begin{center}
\begin{tabular}{lccc} 
Dataset & {\# Points} & {Held-out} & Max. \# States\\
\hline
Synthetic & $5e^3$ & $15\%$ & $9$ \\
Users & $1.4e^4$ & $10$ \%  & $23$ \\
Sensors  &  $1.2e^4$ & $10$ \%  &  $10$\\
Drosophila & $1e^4$  & $15$ \% & $12$\\
\end{tabular}
\end{center}
\label{tbl:datests}
\end{table}

\begin{table*}[t]
\vspace{-5mm}
\caption{Labeling error and predictive log-likelihood for various different datasets.  
\textbf{Key:} iHMM~=~infinite HMM; sub-iHMM~=~a two level hierarchical infinite HMM; siHMM (WoF)~=~feature-independent siHMM; siHMM (WF)~=~feature-based siHMM; 
Synthetic~=~synthetic data segmentation and hidden state inference tasks for which we report the normalized Hamming distance instead of error rate; Users~=~user trace segmentation task; Sensors~=~labeling sensor data task;  Drosophila~=~segmenting fruit fly behavior task.
}
\vspace{-1mm}
\begin{center}
\begin{tabular}{lcccc} 
& \multicolumn{4}{|c}{(Normalized Hamming Distance / Error\% \& Predictive LL)} \\
{Data Set}  & \multicolumn{1}{|c}{iHMM} & {sub-iHMM} & {siHMM (WoF)} & {siHMM (WF)} \\
\hline
Synthetic & $(0.21, -2.07 \times 10^3)$ & $(0.23, -2.24\times 10^3)$ & $(0.15, -2.05\times 10^3)$& $(\textbf{0.13}, -2.19\times 10^3)$\\
Users &  $(30\%, -3.87 \times 10^3)$ &  $(24\%,-2.51 \times 10^3)$ & $(25\%,-3.71 \times 10^3)$ & $(\textbf{16\%}, -3.52 \times 10^3)$ \\
Sensors  & $(27\%, -3.25 \times 10^3)$  & $(22\%, -3.95 \times 10^3)$ & $(\textbf{18\%}, -3.21 \times 10^3)$ & $(34\%, -3.35 \times 10^3)$\\
Drosophila & $(36\%, -6.58 \times 10^4)$ & $(41\%, -6.7 \times 10^4)$ & $(37\%,-7.18 \times 10^4)$ & $(\textbf{34\%},-6.71 \times 10^4)$ \\
\end{tabular}
\end{center}
\vspace{-4mm}
\label{tbl:experiments}
\end{table*}

\subsection{Segmenting user behavior traces} 
\label{sec:userexp}
Modeling user behavior traces is of significant importance in human-computer interaction literature (e.g., \cite{adar2014commandspace, bigham2014human}). Having a good understanding of the tasks done by users can potentially help in designing better work-flows in software applications. It can also help in providing better guidance to users by predicting user intentions. Log files of software applications contain user actions and their corresponding time-stamps; however, it is not clear only from these log files how many different tasks have been done by a user in a single work session. A task consists of multiple actions and each work session consists of multiple tasks. Our two-level hierarchical model can be used for detecting the boundaries between tasks (i.e., segments). We believe that for the software applications in which the tasks are less predictable compared to the actions within each task, our model is a good fit. For instance,  in a photo editing software, the high-level tasks such as adding filters or removing a part of a picture usually do not have a clear order and different users, based on their needs, apply different orderings. However, the actions required to add a filter (e.g., 1- choose a filter, 2- select a part of the photo, and 3- apply the filter) are typically more structured and follow an order. 

We collect log files of users who follow 23 different tutorials in a photo editing software. The dataset is in the form of time-stamp and action; it contains 14000 data points and 59 unique actions in total. We randomly choose a sequence of size 1400 and form a held-out set. We split the sequences into subsequences of size 1000 and apply 100 passes of SVI to the dataset with both feature-based and feature-independent models. 

The possible hyperparameter settings that we consider are $K \in \{60, 80\}$, $\alpha \in \{1, 3\}$, $\gamma \in \{1,3\}$ and finally $\alpha_0 \in \{1, 10\}$, the parameter for the Dirichlet prior, which we place on the parameters of the observation likelihood. We run SVI with 10 different seeds for each setting and report the result for the best setting with the highest VLB in Table~\ref{tbl:experiments}. 

The labels (i.e., the tutorial numbers that the user followed) for each segment of the dataset are available; hence, we can test the siHMM on predicting the labels for each segment. This task is more involved than segmentation, as we also need to group the substates. We use a simple K-means clustering on the empirical transition matrix which is generated from counting the transitions in each inferred segment. This approach works well in practice; however, more sophisticated methods are possible for grouping the substates.  Table~\ref{tbl:experiments} provides the prediction error (computed using the Munkres algorithm) for siHMM and the baselines. The performance of our feature-independent model is significantly better than iHMM and comparable with that of the more flexible (and also computationally intensive) sub-iHMM. Adding the observation feature to the model reduces the error to 16\%. As mentioned in Section \ref{sec:Feature-based model}, in this dataset there are observations that can signal a change-point in the dynamics. 



\subsection{Segmenting fruit fly behavior}
\label{sec:fruitflyexp}
Automating scientific experiments on live animals has attracted significant attention recently (see, for instance, \cite{kain2013leg, wiltschko2015mapping, crall2015beetag, freeman2014mapping} ). With the advent of high throughput and more accurate devices, the need for automatic analysis of large amounts of collected data is felt more today. In neuroscience and biology, a large amount of behavioral data is collected from live animals in order to understand how the brain generates activity and how the underlying mechanisms have evolved \cite{kain2013leg}. Typically the first step in analyzing this data, is finding and categorizing different types of behavior; this step can be done manually by experts but it is time-consuming and sometimes error-prone. 

An automatic framework has been proposed in \cite{kain2013leg} for tracking the leg movements and classifying the behavior of fruit flies. The behavior is recorded by tracking each leg of a fruit fly moving upon a track ball. The collected raw data is the \textit{x} and \textit{y} coordinates of 6 legs and the three rotational components of the rotating ball (i.e., a 15-dimensional vector in real time). After some post-processing and adding some higher-order features (e.g., derivatives of each of the 15 raw data vectors), they expand the dimensions to 45 and apply a KNN classifier to classify each frame as a part of 12 possible behavioral labels. Our goal is to use this dataset and categorize the frames in an unsupervised way. A frequent assumption in the behavioral sciences is that a small set of stereotyped motifs describe most animal activities \cite{berman2014mapping}. In other words, actions within a behavioral segment (e.g., actions required for grooming) should be more structured, compared to the behavioral segments themselves. Given the capabilities of the siHMM, it is a reasonable choice for applying to this dataset.  

The dataset contains 10000 data points; our held-out set is a randomly chosen subsequence with length 1500, and we apply SVI for 100 passes over both the feature-independent and feature-based models. For the observations, we use a multivariate Gaussian likelihood and a conjugate Normal/inverse-Wishart prior. We use the empirical mean and variance of the dataset as the mean and variance of the Gaussian prior. The parameters of the inverse-Wishart are chosen from the possible combinations of $\nu \in \{47, 50\}$ and $\kappa \in \{1, 0.1, 0.01, 0.001\}$. We set the truncation level for the states to 20. Finally, we have the following possible settings for the transition matrix prior $\alpha \in \{1, 10\}$ and $\gamma \in \{1, 10\}$.       

\begin{figure}[bt]
\begin{center}
\includegraphics[width=1.05\columnwidth ]{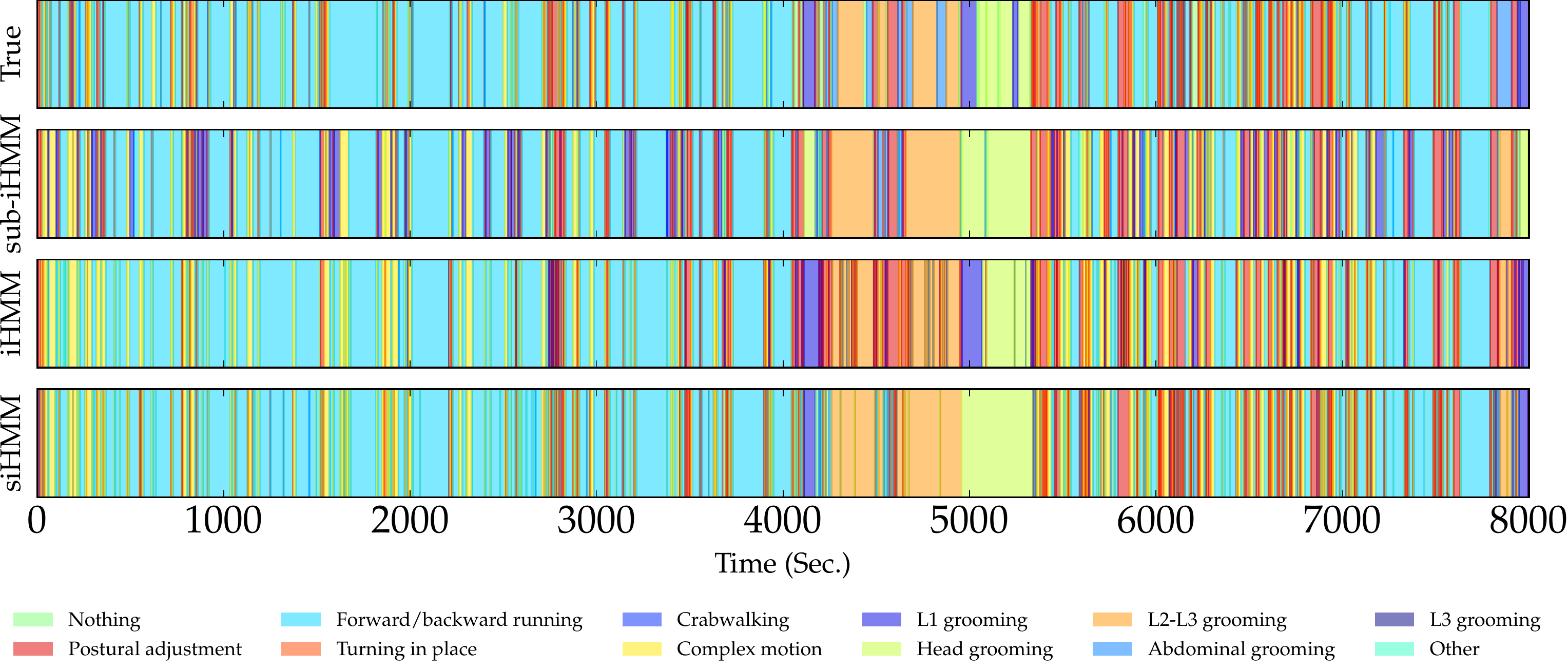} 
\caption{\footnotesize{A sample segmentation from the fruit fly dataset. From top to bottom: True, sub-iHMM, iHMM, siHMM.}}
\label{fig:fruitflydata}
\end{center}
\end{figure}

As in Section~\ref{sec:userexp}, we group the inferred substates with K-means and assign labels to the segments. The results, presented in Table~\ref{tbl:experiments}, show that feature-based siHMM performs on a par with the iHMM and outperforms sub-iHMM by a relative error reduction of 17\%.  This may emphasize the importance of adding data-driven features to the model. Figure~\ref{fig:fruitflydata}, shows a sample of the dataset and its segmentation by different methods. \\


\subsection{Segmenting and classifying human behavior from sensor data}
\label{sec:biosensorexp} 

Through the emergence of pervasive computing and affordable wearable sensors, in-situ measurement of different bio-signals has become possible. This powerful source of data can be utilized for several purposes, including activity recognition and task identification. Toward this goal, an efficient algorithm for analyzing this large amount of data -- which is gathered 24/7 -- is essential. In this section, we use siHMM to model the data collected via Empatica E4 wristband~\cite{E4}, a wearable device that can collect Electrodermal activity (EDA) \cite{boucsein2012electrodermal}, blood volume pulse (BVP), acceleration, and body temperature. EDA refers to changes in electrical properties of the skin caused by sudomotor innervation \cite{boucsein2012electrodermal}. EDA is an indication of physiological or psychological arousal and has been utilized to objectively measure affective phenomena and sleep quality \cite{sano2015discriminating}. 

Segmenting the sensor data can help psychophysiological activity recognition. For instance, it can help in finding stressful periods objectively in order to detect the roots of stress in a person's lifestyle. However, manual labeling for large amounts of user data (days or months) is time-consuming and even invalid if not reported in a timely manner. 

We use a dataset with 12000 time steps, collected from a single user, and model the (normalized) observations (i.e., EDA, BVP and acceleration in 3 dimensions) by a multivariate Gaussian distribution. The hyperparameter setting is similar to that of Section \ref{sec:fruitflyexp}. 

\begin{figure}[bt]
\begin{center}
\includegraphics[width=1.05\columnwidth ]{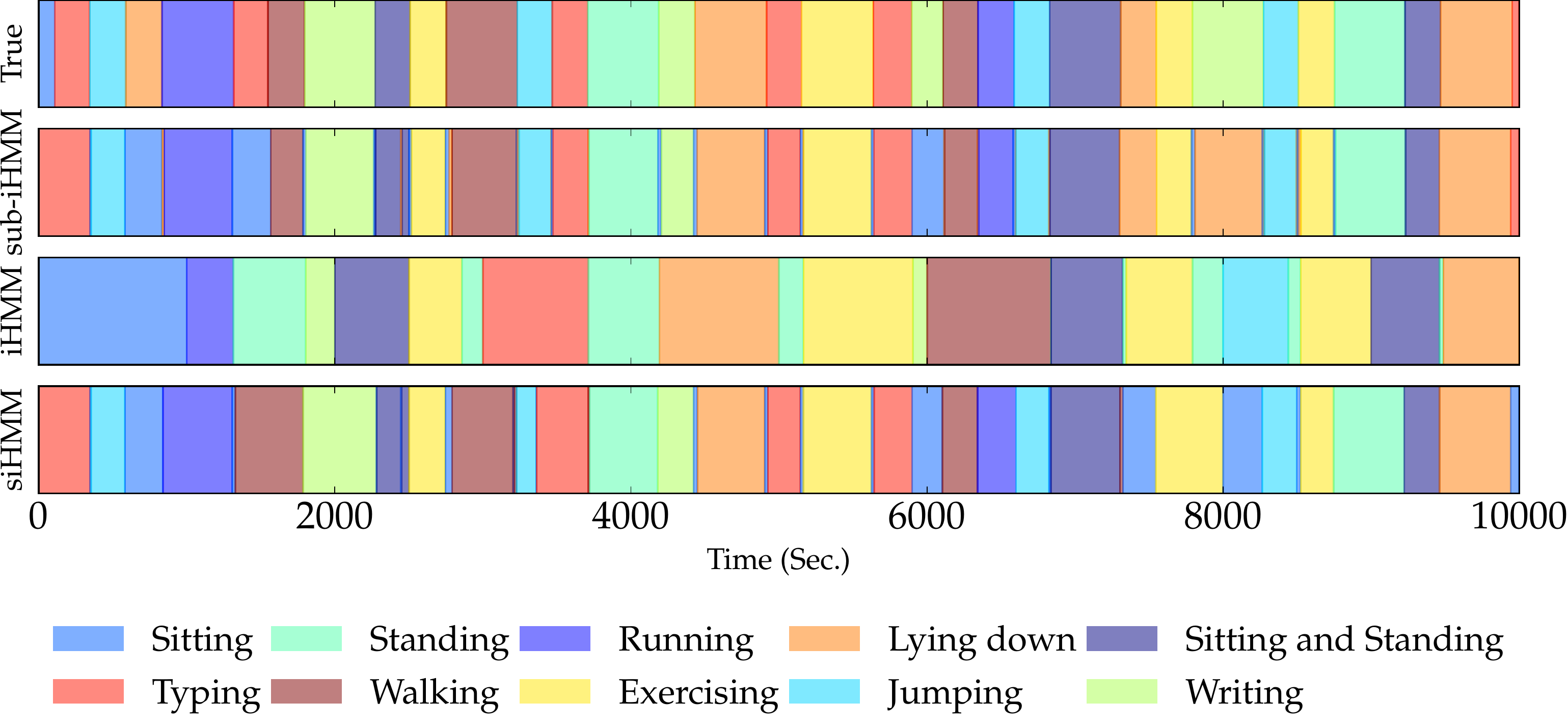} 
\caption{\footnotesize{Segments and labels for data collected from sensors. From top to bottom: True, sub-iHMM, iHMM, siHMM.}}
\label{fig:sensor}
\end{center}
\end{figure}

The labeling error in Table \ref{tbl:experiments} shows that feature-independent siHMM, while performing comparably to sub-iHMM, outperforms iHMM by a relative error reduction of 50\%. Fig. \ref{sec:biosensorexp} demonstrates the inferred segments for siHMM and the baselines. It seems that the single observation feature that we are using for the experiment does not help in this dataset; however, more sophisticated features may help improving the segmentation.


\section{Conclusion}
We proposed a new Bayesian nonparametric model, siHMM, for modeling dynamics at two timescales in time series. Our model is a simple extension to the widely used iHMM and has an efficient inference scheme. Although our model is less flexible than other nonparametric models for hierarchical time series, we showed that it can perform reasonably well in practice. One potential application of our model is using the inferred state-independent transition vector ($\pi_0$) for summarizing a sequence. For instance, in the user behavior analysis, this vector may represent a user fingerprint and users can be grouped based on it. For a better understanding of this feature and the behavior of our model in other applications, a more comprehensive comparison with other models is useful.

\renewcommand*{\bibfont}{\small}
\bibliographystyle{icml2016}
\bibliography{seg-hdp}

\end{document}